%% file: main.tex
\documentclass{edm_article}
\input{macros}

\begin{document}

\title{Evaluating and Optimizing Educational Content with\\ Large Language Model Judgments}

% % Submissions for EDM are double-blind: please do not include any author names or affiliations in the submission. 
% % Anonymous authors:
% \numberofauthors{1}
% \author{
% Anonymous\\
%        \affaddr{Anonymous Institution}\\
%        \email{anonymous@anonymous.edu}
% }

%An example of how to include
% multiple authors is below for after the paper has been accepted.


\numberofauthors{3}
\author{
\alignauthor
		Joy He-Yueya\\
       \affaddr{Stanford University}\\
       \email{heyueya@cs.stanford.edu}
\alignauthor
		Noah D. Goodman\\
       \affaddr{Stanford University}\\
       \email{ngoodman@stanford.edu}
\alignauthor
		Emma Brunskill\\
       \affaddr{Stanford University}\\
       \email{ebrun@cs.stanford.edu}
}

\maketitle

\begin{abstract}
Creating effective educational materials generally requires expensive and time-consuming studies of student learning outcomes. To overcome this barrier, one idea is to build computational models of student learning and use them to optimize instructional materials. However, it is difficult to model the cognitive processes of learning dynamics. We propose an alternative approach that uses Language Models (LMs) as educational experts to assess the impact of various instructions on learning outcomes. Specifically, we use GPT-3.5 to evaluate the overall effect of instructional materials on different student groups and find that it can replicate well-established educational findings such as the Expertise Reversal Effect and the Variability Effect. This demonstrates the potential of LMs as reliable evaluators of educational content. Building on this insight, we introduce an instruction optimization approach in which one LM generates instructional materials using the judgments of another LM as a reward function. We apply this approach to create math word problem worksheets aimed at maximizing student learning gains. Human teachers' evaluations of these LM-generated worksheets show a significant alignment between the LM judgments and human teacher preferences. We conclude by discussing potential divergences between human and LM opinions and the resulting pitfalls of automating instructional design. \footnote{Our prompts are publicly available at \href{https://github.com/StanfordAI4HI/ed-expert-simulator}{https://github.com/StanfordAI4HI/ed-expert-simulator}.}
\end{abstract}


\input{sections/1_introduction}
\input{sections/2_related_work}
\input{sections/3_replicate_experiments}

\input{sections/4_optimization}
\input{sections/5_conclusion}

\section{Acknowledgements}
We would like to thank Dongwei Jiang, Rose E. Wang, Allen Nie, and Shubhra Mishra for their feedback on our work. This work was supported by the Junglee Corporation Stanford Graduate Fellowship and a Stanford Hoffman-Yee grant.

\bibliographystyle{abbrv}
\bibliography{references}  % sigproc.bib is the name of the Bibliography in this case
%\balancecolumns

\clearpage
\onecolumn
\appendix
\input{sections/appendix}

\balancecolumns
% That's all folks!
\end{document}

%% file: macros.tex
\usepackage{graphicx}
\usepackage{booktabs} % for professional tables
\usepackage{stackengine} % for stackunder
\usepackage{float} % for [H]
\usepackage{wrapfig} % wrapfigure
\usepackage{color}
\usepackage{algorithm}
\usepackage{algpseudocode}

\usepackage{xcolor} % Add color support
\definecolor{cbBlue}{RGB}{0, 114, 178} % Blue
\definecolor{cbOrange}{RGB}{230, 159, 0} % Orange
\definecolor{cbGreen}{RGB}{0, 158, 115} % Bluish Green
\definecolor{tableauBlue}{RGB}{31,119,180}
\definecolor{tableauOrange}{RGB}{255,127,14}
\definecolor{tableauGreen}{RGB}{44,160,44}
\definecolor{tableauRed}{RGB}{214,39,40}
\definecolor{tableauPurple}{RGB}{148,103,189}
\definecolor{tableauBrown}{RGB}{140,86,75}
\definecolor{tableauPink}{RGB}{227,119,194}
\definecolor{tableauGray}{RGB}{127,127,127}
\definecolor{tableauOlive}{RGB}{188,189,34}
\definecolor{tableauCyan}{RGB}{23,190,207}

\usepackage{hyperref}

%% file: sections/1_introduction.tex
\section{Introduction}
Instructional design, the process of creating educational materials or experiences such as textbooks and courses, is a critical component in advancing education \cite{smith2004instructional}. A significant challenge in instructional design is the need for extensive studies involving real students to evaluate the effectiveness of these instructional materials \cite{komoski1974eric, nieveen2013formative}. This traditional method is expensive, time-consuming, and fraught with logistical and ethical challenges, making it difficult to quickly innovate and implement new teaching strategies. Recent studies have explored the use of Language Models (LMs) to simulate students' interactions with educational content \cite{markel2023gpteach, jinxin2023cgmi}, offering a less expensive alternative. However, our initial investigations reveal that the LMs (at the time) struggled to model the \textit{dynamics} of student learning, often failing to maintain a consistent level of knowledge when simulating students' responses before and after learning interventions (see Appendix \ref{appendix:student_simulator_challenge}). 

In light of these challenges, our work explores the use of LMs, such as GPT-3.5 and GPT-4 \cite{achiam2023gpt}, to evaluate and optimize educational materials. Unlike previous attempts to directly simulate student learning \cite{markel2023gpteach, jinxin2023cgmi, nguyen2023large, opedal2024language, mertz1997using}, we leverage the advanced reasoning capabilities and pedagogical knowledge of LMs and position LMs as educational experts who can assess and enhance the effectiveness of instructional content. 

To validate the potential of LMs in simulating educational experts, we use GPT-3.5 to assess the overall effect of instructional materials on different student groups and find that the judgments of GPT-3.5 can replicate two well-known findings in educational psychology: the Expertise Reversal Effect \cite{kalyuga2003expertise, tuovinen1999comparison, kalyuga2004measuring, kalyuga2007expertise} and the Variability Effect \cite{paas1994variability, sweller2019cognitive}. These results suggest that LMs have the potential to act as coherent evaluators of instructions, offering insights consistent with those obtained from human subjects research.

\begin{figure*}[ht]
\centering
\includegraphics[width=0.93\textwidth]{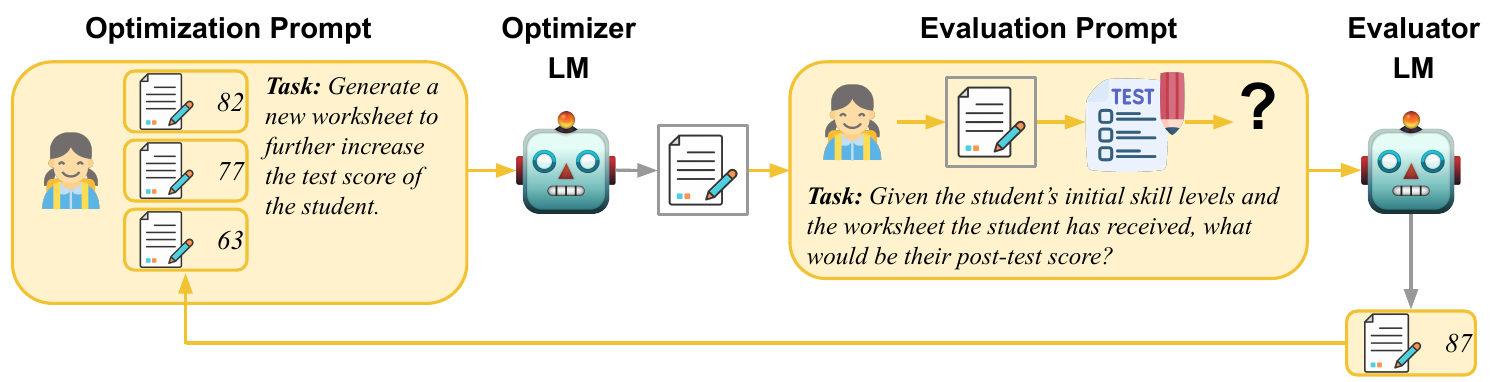}
\caption{We use an optimizer LM to generate new educational materials (e.g., worksheets) and an evaluator LM to judge the effectiveness of these materials by predicting students' post-test performance. In this example, the optimizer LM generates a new worksheet based on the student information and previous worksheet-score pairs, then the evaluator LM predicts the student's post-test score given the new worksheet. The new worksheet-score pair serves as feedback for the optimizer LM to refine worksheets.}
\label{fig:framework}
\Description{We use an optimizer LM to generate new educational materials (e.g., worksheets) and an evaluator LM to judge the effectiveness of these materials by predicting students' post-test performance. In this example, the optimizer LM generates a new worksheet based on the student information and previous worksheet-score pairs, then the evaluator LM predicts the student's post-test score given the new worksheet. The new worksheet-score pair serves as feedback for the optimizer LM to refine worksheets.}
\end{figure*}

Building on the insight that LMs can, to some extent, mimic educational experts, along with prior work demonstrating LMs' capability for iterative improvement \cite{madaan2023self, bai2022constitutional, zelikman2023self, yang2023large, pryzant2023automatic}, we propose an instruction optimization approach (Figure \ref{fig:framework}). In this approach, one LM (the \textit{optimizer}) generates new instructional materials, and another LM (the \textit{evaluator}) evaluates these materials by predicting students' learning outcomes (e.g., post-test scores). We apply this approach to optimize math word problem worksheets, aiming to maximize students' post-test scores. External assessments like post-tests are not perfect, but they are an important tool in many educational systems to help shed insight into student progress. In each optimization step, we prompt the optimizer LM to generate new worksheets based on a list of previously generated worksheets with their corresponding post-test scores, then the new worksheets are scored by the evaluator LM. We show that the optimizer LM can refine worksheets based on the judgments of the evaluator LM. We also ask human teachers to compare worksheets generated from different stages of optimization and find a significant correlation between the evaluator LM's judgments and human preferences. This highlights the potential of LMs in informing the design of real-life experiments and reducing the number of costly experiments in education. However, human teachers sometimes cannot distinguish worksheets that LMs perceive as different, which suggests the necessity for further investigations to ensure LMs effectively complement traditional educational research methodologies.

Our paper makes the following contributions:
\vspace{-0.4cm}
\begin{itemize}
    \item We demonstrate the potential of LMs to serve as reliable evaluators of educational content by replicating two well-established educational findings: the Expertise Reversal Effect and the Variability Effect.
    \item We introduce an instruction optimization approach using the LM judgments as a reward function and demonstrate the feasibility of using LMs to iteratively improve educational materials, focusing on the domain of math word problem solving.
    \item We recruit human teachers to evaluate LM-generated math word problem worksheets from different stages of optimization and find a significant alignment between human teacher preferences and LM judgments, highlighting a promising application of LMs in informing the design of costly human subjects experiments in education.
\end{itemize}
\vspace{-0.3cm}

After describing these results, we conclude the paper with a discussion of open issues and future directions. 

%% file: sections/2_related_work.tex
\section{Related work}
There is an extensive and growing literature on simulating human behaviors and LM-based tools for learning and instructional design.

\subsection{Simulating students for education}
Long before LM-based tools, researchers explored methods to build simulated students (usually machine learning systems whose behavior is consistent with data from human students) and applications of such simulated students in education \cite{vanlehn1994applications, anderson1995cognitive, greer2013student, mertz1997using, matsuda2007evaluating}. For example, \cite{vanlehn1994applications} discusses how teachers can develop and practice their tutoring strategies by teaching simulated students, as demonstrated in recent work that develops an interactive chat-based tool that allows teachers to practice with LM-simulated students \cite{markel2023gpteach}. Another promising application of student simulators considered in \cite{vanlehn1994applications} is to enable collaborative learning where students can work collaboratively with a simulated peer or by teaching a less knowledgeable simulated student \cite{matsuda2010learning, ma2023hypocompass, schmucker2023ruffle, jin2023teach}. 

Our work is closely related to prior studies on simulation-based evaluations of instructional design \cite{ohlsson1992cognitive, vanlehn1994applications, mertz1997using}. Running empirical experiments with real students to test different instructions, often called formative evaluations \cite{nieveen2013formative}, can help instructional designers pilot-test and improve their designs. Since formative evaluations with real students are expensive, it is of interest to build simulated students who can provide cheap, fast evaluations. \cite{ohlsson1992cognitive} introduces Heuristic Searcher, a system that can be used as a simulated student who learns arithmetic skills from instructions encoded as formal constraints that describe erroneous states. They show that simulations can reveal interactions of instruction with students' prior knowledge. Similarly, \cite{mertz1997using} uses a cognitive model as a simulated student to help design instructions for training circuit board
assemblers. However, these systems are limited by their inability to process instructions in natural language or adapt to broader, less structured knowledge domains. Although advancements in LMs \cite{achiam2023gpt, bubeck2023sparks} offer opportunities to simulate students who can interact with educational content in natural language, recent attempts show that LM-simulated students can at times produce responses that are unrealistic or overly advanced, particularly in situations where such capability is not expected \cite{markel2023gpteach, aher2023using}. Therefore, instead of focusing on simulating students who can interact with instructions, we aim to simulate educational experts capable of evaluating the quality of those instructions.

\subsection{LMs as educational experts and tutors}
Recent advancements in LMs \cite{achiam2023gpt, bubeck2023sparks} have sparked a new wave of work that explores ways in which LMs can improve education. Some studies have considered using LMs for educational content development \cite{sarsa2022automatic, schmucker2023ruffle, denny2023can, wang2022towards, zelikman2023generating}. \cite{sarsa2022automatic} uses LMs to produce educational resources (e.g., code examples and explanations) for an introductory programming course. They find that students perceive the quality of LM-generated resources to be similar to materials produced by their peers. \cite{wang2023chatgpt} explores the ability of LMs to become teacher coaches who can provide teachers with pedagogical suggestions. There have been many papers that focus on using LMs to provide automated feedback and explanations to students learning math \cite{pardos2023learning, wang2023step, kumar2023math, prihar2023comparing}, programming \cite{pankiewicz2023large, phung2023automating}, and physics \cite{riabko2024physics}. None of these studies considers optimizing educational content for a given student.

\subsection{LMs for simulating humans}
Our work is loosely related to recent work on using LMs to simulate different aspects of human behavior. For instance, researchers have used LMs to replicate results from social science experiments and public opinion surveys \cite{argyle2023out, aher2023using, horton2023large}. The ability of LMs to mimic human behaviors opens up exciting possibilities in areas such as education \cite{markel2023gpteach}, product design \cite{park2022social, park2023generative}, and skill training \cite{shaikh2023rehearsal, liu2023improving}.

%% file: sections/3_replicate_experiments.tex
\section{Instruction Evaluation}

To do instructional design without running experiments with real students, we introduce a new type of evaluation for assessing the impact of instructions on student learning gains. 

\subsection{LM-based simulated expert evaluation}

Traditionally, developing effective instructional materials involves administering pre-tests and post-tests to numerous students under different experimental conditions. However, this type of evaluation is expensive and time-consuming \cite{vanlehn1994applications, komoski1974eric}. Another option is to have pedagogical experts create content based on their (often implicit) assessment of how well it will support student learning \cite{williams2011instructional}. Such human experts can also judge whether particular instructions are likely to be more or less effective.

Inspired by this, we use an LM to simulate a human pedagogical expert evaluation of a particular instructional content that can be described in text. More precisely, we task an LM to take in information about a student's relevant background  (e.g., the student's prior knowledge), a particular set of instructional content that would be provided to the student, and predict the student's subsequent performance on particular test questions. We outline this procedure in Alg.~\ref{alg:evaluator}.  We call this evaluation a Simulated Expert Evaluation (SEE). In general, we are interested in understanding whether a particular instructional artifact is effective for a group of students, such as eighth graders doing algebra, or fifth graders struggling with fractions. To do this evaluation, we will have a simulated expert assess the impact on many different students with different prior knowledge or background, to get an aggregate estimate of the impact of a particular instruction.

\begin{algorithm}
 \caption{Simulated Expert Evaluation (SEE)}
  \begin{algorithmic}
    \State \textbf{Input:} Evaluator $M_e$, \textit{student\_persona}, \textit{instruction}, \textit{eval\_task}, test questions $x_1,...,x_T$
    \State Initialize $m_e$ as an instance of $M_e$
    \State Set $p_1 \gets \textit{student\_persona} || \textit{instruction} || x_1 || \textit{eval\_task}$
    \State Give $p_1$ as input to $m_e$ and receive output $o_1$
    \For{$i = 2,3,\dots T$}
        \State Set $p_i \gets x_i || \textit{eval\_task}$
        \State Give $p_i$ as input to $m_e$ and receive output $o_i$
    \EndFor       
    \State Extract $s_1,...,s_T$ from $o_1,...,o_T$
    \State Set $s \gets \frac{\sum_i^T s_i}{T}$
    \State \Return $s$
  \end{algorithmic}
  \label{alg:evaluator}
  \end{algorithm}

\begin{figure*}[ht]
\noindent\fbox{
\parbox{0.975\textwidth}{

\color{tableauBlue}
Here is an 8th-grade student with the following skill levels (each skill is rated on a scale from 1 to 5):
\begin{enumerate}
    \item Being able to set up systems of equations given a word problem: \{level1\}
    \item Being able to solve systems of equations: \{level2\}
\end{enumerate}

\quad  

\color{black}

Here's the instruction that the student receives. The student is asked to study a problem and its solution. Here's the problem:

\quad  

A brownie recipe is asking for 350 grams of sugar, and a pound cake recipe requires 270 more grams of sugar than a brownie recipe. How many grams of sugar are needed for the pound cake? 

Here's its solution:

Step 1: Identify the amount of sugar needed for the brownie recipe, which is 350 grams.

Step 2: Understand that the pound cake recipe requires 270 more grams of sugar than the brownie recipe.

Step 3: Add the additional 270 grams of sugar to the 350 grams required for the brownie recipe.

Step 4: The total amount of sugar needed for the pound cake recipe is 350 grams + 270 grams = 620 grams.

\quad  

The student is then asked to work on the following problem on their own:

The size of a compressed file is 1.74 MiB, while the size of the original uncompressed file is 5.5 times greater. What is the size of the uncompressed file, in MiB?

\quad

(… more instructions …)

\color{tableauOrange}

\quad 

Now the student is asked to work on the following problem on a test: 

\quad   

Alyssa is twelve years older than her sister, Bethany. The sum of their ages is forty-four. Find Alyssa's age.

\quad

\color{tableauGreen}
Given the student's initial skill levels and the instruction the student has received, what's the probability that the student can solve the problem correctly? Explain your reasoning and give a single number between 0 and 100 in square brackets.

}  
}
\caption{The prompt for predicting the post-test score in a simulated expert evaluation. The \textcolor{tableauBlue}{blue} text is the student persona. The \textcolor{black}{black} text is the instruction given to the student. The \textcolor{tableauOrange}{orange} text is a problem on the post-test. The \textcolor{tableauGreen}{green} text describes the evaluation task for the LM.}
\label{fig:see_post_test_prompt}
\Description{The prompt for predicting the post-test score in a simulated expert evaluation. The \textcolor{tableauBlue}{blue} text is the student persona. The \textcolor{black}{black} text is the instruction given to the student. The \textcolor{tableauOrange}{orange} text is a problem on the post-test. The \textcolor{tableauGreen}{green} text describes the evaluation task for the LM.}
\end{figure*}

\begin{figure}[ht]
\noindent\fbox{
\parbox{0.95\columnwidth}{

\color{tableauOrange}

\quad 

Now the student is asked to work on the following problem on a test: 

\quad   

\{problem\}

\quad

\color{tableauGreen}
Given the student's initial skill levels and the instruction the student has received, what's the probability that the student can solve the problem correctly? Explain your reasoning and give a single number between 0 and 100 in square brackets.

}  
}
\caption{The \textcolor{tableauOrange}{orange} text is the test question. The \textcolor{tableauGreen}{green} text describes the evaluation task for the LM.}
\label{fig:see_individual_test_prompt}
\Description{The \textcolor{tableauOrange}{orange} text is the test question. The \textcolor{tableauGreen}{green} text describes the evaluation task for the LM.}
\end{figure}

More precisely, the input of an SEE are an evaluator LM $M_e$ (for example, ChatGPT/Claude/Gemini/etc), a text description of a student persona, a text instruction, an evaluation task description, and $T$ test questions $x_1,...,x_T$. In an SEE, we first initialize an evaluator LM instance $m_e$. We create a list of prompts $\{p_1, p_2, ..., p_T\}$ and sequentially give these prompts as input to $m_e$. The initial prompt $p_1$ (see an example in Figure \ref{fig:see_post_test_prompt}) includes the student persona that represents the student's proficiency in relevant skills, the instruction, the first problem on the post-test $x_1$, and the task for the LM (which is to predict the probability that the student can solve the problem correctly). For example, in the figure we show a case where a student is learning to solve systems of equations. The student starts with some skill levels, and then the student is provided with a worked example, and then given a problem to work out a solution to a new problem, along with other activities (omitted for space). As shown in \textcolor{tableauOrange}{orange}, after such ``simulated'' instruction, the student will be asked a test question (in \textcolor{tableauOrange}{orange})-- here a word problem requiring the student to set up and solve two equations to work out the age of a sister. In \textcolor{tableauGreen}{green} the LM $m_e$ will be instructed to predict the performance on this question of the student who start from this initial background knowledge and completed the provided activities (worked example etc). More formally, the LM evaluator $m_e$ takes in this $p_1$ (as shown in Figure \ref{fig:see_post_test_prompt}) and outputs a response $o_1$. We then feed the LM $m_e$ the subsequent prompts $p_2, ..., p_T$, where each prompt $p_i$ includes an additional test problem $x_i$ and the task description (see an example in Figure \ref{fig:see_individual_test_prompt}). Since we give all the prompts to the same LM instance $m_e$, $m_e$ generates $o_i$ in the context of $\{p_1, o_1, p_2, o_2, ..., p_{i}\}$. Essentially this repeated interaction just allows us to have the LM $m_e$ evaluate how the student persona would do on the post test, given the provided instruction.\footnote{We also explored providing all the test questions and then asking the LM to predict the probability of the student getting each right, but we found that asking the LM to predict one by one was more effective. } After we collect $o_1,...,o_T$, we extract the probability of success (in percentage) for each test problem $s_1, ..., s_T$ and calculate the post-test score $s$ as the average probability of success:
\begin{equation}
    s = \frac{\sum_i^T s_i}{T}
\end{equation}
We will later discuss the specific student personas (student background description) we use.

A simulated educational expert offers many benefits. For example, we can easily get a simulated expert assessment of the impact of different instructional content on the same student by altering the instructional content in the prompt. Additionally, by setting the instructional content to an empty string (see Figure \ref{fig:see_pre_test_prompt} in Appendix), we ask the simulated expert to estimate a student's pre-test score, allowing for the use of identical questions in both pre-tests and post-tests. This approach allows us to estimate learning gains while avoiding biases associated with memorization.

Importantly, using an LM as a simulated educational expert is different from using an LM to directly simulate a given student. Simulating a student involves modeling the student's problem-solving process, while an SEE only predicts the post-test scores without actually simulating the student's responses. In our early explorations, we found that the LMs we were using were poor simulators of student learning processes (see Appendix \ref{appendix:student_simulator_challenge}). For example, while an LM might successfully offer an answer as to what a fifth grader might respond to an algebra question, if we then provided some pedagogical material and asked the LM to directly simulate what the student would now respond to a test question, the LM frequently could ``forget'' its (the fifth graders') limited knowledge and answer the question perfectly. In other words, the LMs (at the time) had a poor model of the \textit{dynamics} of student learning. To emphasize the distinction using our example, a simulated educational expert LM will predict the chance the student will correctly answer the sister age question (e.g., 70\% chance the student will get it right) whereas an LM-as-student must generate the actual answer (e.g., ``Alyssa is 32'').

We now evaluate to what extent a given LM, such as GPT-3.5, can act as a simulated educational expert and predict the impact of instructions on student learning outcomes. In particular, such a simulated educational expert should align with prior known results about student learning from different types of instructional materials. Our primary interest is to consider if we can use LMs as simulated educational experts to evaluate and optimize new instructional content. Therefore, our focus here is not a comprehensive evaluation of the ability of LMs to match known results in the learning sciences, but to do a small set of basic tests to make sure there are reasons to believe that an LM might be a good simulated educational expert. 

In particular, we run simulated expert evaluations (SEEs) over a population of student personas to see if the judgments of our simulated educational expert LMs can replicate two well-known instructional effects: the Expertise Reversal Effect \cite{kalyuga2003expertise, tuovinen1999comparison, kalyuga2004measuring, kalyuga2007expertise} and the Variability Effect \cite{paas1994variability, sweller2019cognitive}. We selected these two phenomena in part because they are compatible with text-only prompts, though in the future we hope to extend LM-based simulated educational experts to multi-modal instructional inputs.

\subsection{Replicating prior studies using SEEs}
\label{sec:replicating_prior_studies}

Replicating previous experiments involves: 1) creating participant personas that include relevant covariates; 2) allocating participants to different experimental conditions; 3) creating instructional materials customized for each experimental group and implementing the intervention; and 4) conducting post-tests to evaluate the impact of the intervention. Optionally, conducting pre-tests to determine baseline measures could also be included.

Since prior studies did not publish their experimental materials, we create new student personas and instructional materials, focusing on the domain of math word problem solving. We create $n=120$ student personas using a fixed template (see the \textcolor{tableauBlue}{blue} text in Figure \ref{fig:see_post_test_prompt}) that describes their initial proficiency in the relevant math skills: the ability to set up systems of equations and the ability to solve systems of equations. The student persona vary only in skill levels, \textit{level1} and \textit{level2}, for which we randomly assigned integers between 1 and 5 (inclusive) to simulate diverse student abilities. We define the experimental groups and assign student personas to these groups using the same procedures as prior studies. We simulate the pre-test stage by running one SEE (see Alg.~\ref{alg:evaluator}) for each student persona. The instruction for all these SEEs is an empty string because the students have not received any instruction yet (see Figure \ref{fig:see_pre_test_prompt} in Appendix). To implement the intervention and collect students' post-test scores, we run multiple SEEs (one for each student persona) with specific instructions tailored to their assigned experimental group. After collecting post-test scores, we compare the overall effect of various instructions on different experimental groups with results from prior studies. We describe the two phenomena we replicate below.

\subsection{Expertise Reversal Effect}

The Expertise Reversal Effect, introduced by \cite{kalyuga2003expertise}, suggests that the effectiveness of instructional strategies changes as a learner's knowledge and skills develop. In the early stages of learning, beginners often benefit from structured guidance, which compensates for their limited background knowledge and helps manage cognitive load. However, as learners gain proficiency, the same strategies that were once helpful can become superfluous or even hinder learning by overloading the cognitive system. For learners who have achieved a higher level of expertise, minimal guidance is preferable because they have built up a sufficient framework of knowledge that allows for efficient organization and processing of new information \cite{kalyuga2007expertise}.

\subsubsection{Prior real-life experiments}
In Experiment 3 of \cite{kalyuga2004measuring}, they replicate the Expertise Reversal Effect in the domain of coordinate geometry problem solving. Their experiment consists of three stages: pre-test, instruction, and post-test. 

During the first stage, they administer a pre-test to all participants (42 Year 9 students from a Sydney Catholic girls’ school). Based on the pre-test scores, they divide participants into two groups: more knowledgeable learners (upper median group) and less knowledgeable learners (lower median group). They then randomly allocate students in each of these two groups to two subgroups: one receives practice-based instruction, and the other receives worked-example-based instruction. This leads to four experimental groups: 1) Low-knowledge/practice, 2) Low-knowledge/worked example, 3) High-knowledge/practice, and 4) High-knowledge/ worked example.

In the second stage, participants in the practice conditions are given 8 problems (numbered from 1 to 8) to solve on their own. Participants in the worked-example conditions are given the same set of 8 problems in the same order but with fully worked-out step-by-step solutions for the problems with odd numbers. 

In the final stage, they ask participants to take a post-test. They find that for less knowledgeable (low-expertise) learners, the worked-example group performs significantly better than the practice group on the post-test. For more knowledgeable learners (high-expertise), there is no significant difference between the worked-example group and the practice group. Similarly, \cite{tuovinen1999comparison} uncovers the Expertise Reversal Effect in their experiment teaching students how to use a database program (see their Figure 1). 

\subsubsection{Replication}
\label{sec:ere_replication}

Following the procedures used in \cite{kalyuga2004measuring}, we run SEEs in a different domain: math word problems involving systems of equations. We create $120$ student personas that describe their initial proficiency in relevant math skills (see Section \ref{sec:replicating_prior_studies}). Based on the sum of skill levels for the two skills, we divide these student personas into two groups: high-expertise learners (upper-median group) and low-expertise learners (lower-median group)\footnote{In our SEEs, we do not need to divide students into low-expertise learners and high-expertise learners based on their pre-test scores since we know all students' latent skill levels.}. For each group, we randomly assign half of the students to the practice condition and the other half to the worked-example condition, which leads to four experimental groups:
\vspace{-0.4cm}
\begin{enumerate}
    \item low-expertise/practice.
    \vspace{-0.2cm}
    \item low-expertise/worked example.
    \vspace{-0.2cm}
    \item high-expertise/practice.
    \vspace{-0.2cm}
    \item high-expertise/worked example.
\end{enumerate}
\vspace{-0.3cm}

We create the two types of instructions (practice and worked example), the pre-test, and the post-test, using math word problems from the \textsc{Algebra} dataset \cite{he2023solving}. We randomly select 8 problems for the two instructions and a different set of 8 problems $\{x_1,...,x_8\}$ for the pre-test and post-test. The practice instruction contains 8 word problems with no solutions, while the worked-example instruction includes the same set of 8 problems with step-by-step solutions for the first, third, fifth, and seventh problems. Detailed instructions are available at \href{https://github.com/StanfordAI4HI/ed-expert-simulator}{https://github.com/StanfordAI4HI/ed-expert-simulator}.

We simulate the pre-tests and post-tests as described in Section \ref{sec:replicating_prior_studies}. Figure \ref{fig:see_pre_test_prompt} in the Appendix shows a prompt for the pre-test. Figure \ref{fig:see_post_test_prompt} shows a prompt for predicting the post-test scores for the worked-example group. We use gpt-3.5-turbo-16k as the evaluator LM for all SEEs, with the temperature parameter set to $0$ for more consistent output.

\subsubsection{Results}
\label{sec:ere_results}

\begin{figure}
\centering
\includegraphics[width=\columnwidth]{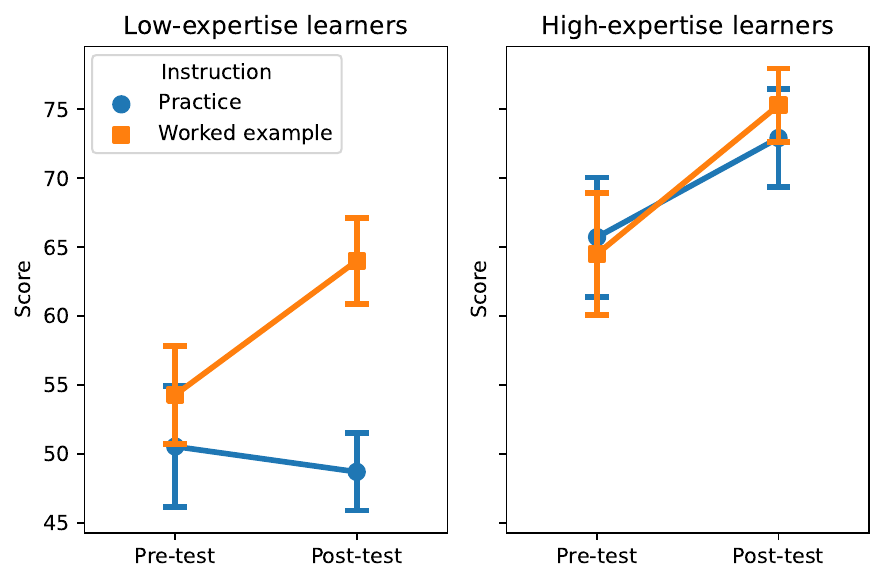}
\caption{The LM-based simulated expert evaluations can replicate the Expertise Reversal Effect. According to the LM judgments, low-expertise learners benefit more from worked examples than practice, but there is no significant difference between practice and worked examples for high-expertise learners. Circles and squares indicate the average test score for each group. The error bars show the standard error.}
\label{fig:ere_results}
\Description{The figure presents two line graphs comparing the scores of low-expertise and high-expertise learners before and after an intervention, which is either a practice or worked-example instruction.}
\end{figure}

We find that LM-based SEEs can accurately replicate the Expertise Reversal Effect, as shown in our comparison of pre-test and post-test scores across all participants (see Figure \ref{fig:ere_results}). Among low-expertise learners, the worked-example group significantly outperforms the practice group on the post-test. This performance difference is not observed among high-expertise learners, where both practice and worked-example groups show similar outcomes.

This finding might seem expected, considering the LM has likely been trained on data that includes descriptions of these prior studies. However, the ability of the LM to make these accurate judgments is particularly noteworthy for two reasons. First, our evaluation methods do not explicitly hint at those prior studies (for instance, we do not specify whether the instruction is practice-based or involves worked examples). Second, we develop and use new instructional materials in a domain different from those explored in prior studies. This highlights the LM's ability to generalize its learning to novel contexts.  

\subsection{Variability Effect}

The Variability Effect \cite{paas1994variability, sweller2019cognitive} suggests that introducing a variety of instructional examples is not always beneficial for students. Variability in problem situations can enhance student learning only if students have sufficient working memory resources to handle the added cognitive load.

\subsubsection{Prior real-life experiment }
The experiment in \cite{paas1994variability} demonstrates the Variability Effect in the domain of geometrical problem-solving. The researchers divide participants (60 students from fourth-year classes of a secondary technical school) into four experimental groups:
\vspace{-0.6cm}
\begin{enumerate}
    \item low-variability/practice.
    \vspace{-0.2cm}
    \item low-variability/worked example.
    \vspace{-0.2cm}
    \item high-variability/practice.
    \vspace{-0.2cm}
    \item high-variability/worked example.
\end{enumerate}
\vspace{-0.3cm}

Participants in the practice conditions receive a set of six problems. Participants in the worked-example conditions receive the same problems with their step-by-step solutions. The problems with odd numbers (first, third, and fifth) are identical for the low-variability and high-variability conditions. The problems with even numbers in the low-variability conditions have the same problem formats as the problems with odd numbers, but with different values. The problems with even numbers in the high-variability conditions have different values and problem formats than the problems with odd numbers. After receiving the intervention, participants take a post-test consisting of six problems. 

The results show that introducing a variety of examples can significantly enhance learning and the ability to apply knowledge, especially when the cognitive load is low, such as when students learn from worked examples. However, in situations where the cognitive load is already high, like when students solve practice problems independently, there is no effect of variability because too much variability can overload a learner's memory capacity. This finding highlights the balance needed between introducing diversity in learning materials and managing cognitive load for optimal learning outcomes.

\subsubsection{Replication}
We run SEEs in the domain of math word problems involving systems of equations, which is the same as Section \ref{sec:ere_replication}. We create $n=120$ student personas that describe their initial proficiency in relevant math skills (see Section \ref{sec:replicating_prior_studies}). Following \cite{paas1994variability}, we randomly divide these student personas into four experimental groups of equal size: 
\vspace{-0.4cm}
\begin{enumerate}
    \item low-variability/practice.
    \vspace{-0.2cm}
    \item low-variability/worked example.
    \vspace{-0.2cm}
    \item high-variability/practice.
    \vspace{-0.2cm}
    \item high-variability/worked example.
\end{enumerate}
\vspace{-0.3cm}

We create four types of instructions (one for each experimental group) and a post-test, using math word problems from the \textsc{Algebra} dataset \cite{he2023solving}. We randomly select a set of 6 problems $\{x_1,...,x_6\}$ for the post-test, which is identical in all conditions. In the instructions for both low-variability and high-variability conditions, the first, third, and fifth problems are identical. In the low-variability conditions, the second, fourth, and sixth problems have the same format as the odd-numbered problems but use different values. In the high-variability conditions, the even-numbered problems differ from the odd-numbered ones in both values and formats. In practice conditions, only problems are presented. In worked-example conditions, all problems and step-by-step solutions are presented. We use gpt-3.5-turbo-16k as the evaluator LM for all SEEs, with the
temperature parameter set to 0 for more consistent output.

\subsubsection{Results}

\begin{figure}[h!]
    \centering
    \includegraphics[width=0.945\columnwidth]{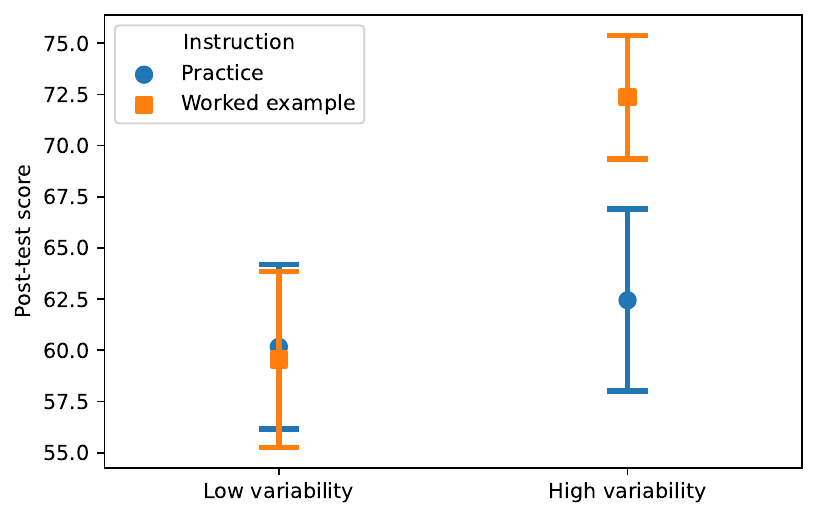}
    \caption{The LM-based simulated expert evaluations can replicate the Variability Effect. According to the LM judgments, high variability in problems significantly enhances performance in worked-example conditions, but there is no effect of variability in practice conditions. Circles and squares indicate the average test score for each group. The error bars show the standard error.}
    \label{fig:var_results}
    \Description{The vertical axis represents post-test scores, ranging from 55.0 to 75.0.}
\end{figure}

\begin{figure}[h!]
\centering
\includegraphics[width=0.945\columnwidth]{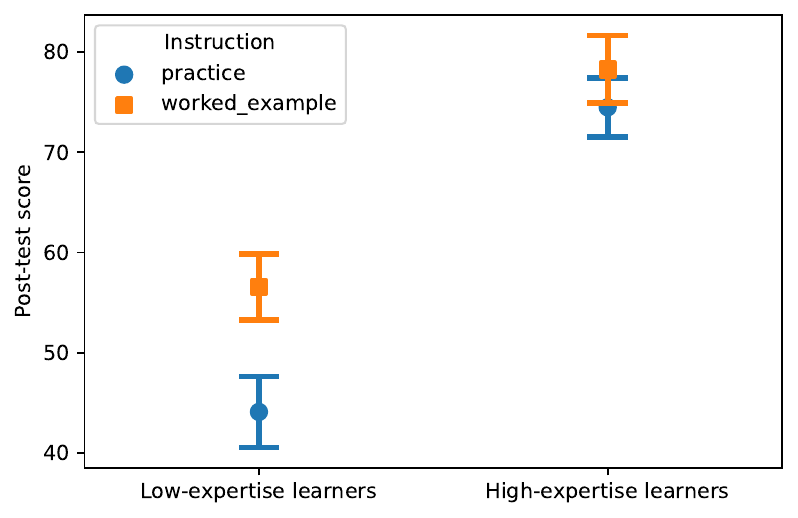}
\caption{The simulated expert evaluations designed for replicating the Variability Effect can also replicate the Expertise Reversal Effect, demonstrating the consistency of LM judgments. Among the low-expertise learners, the worked-example group performs significantly better than the practice group on the post-test. Among the high-expertise learners, there is no significant performance difference between the practice and worked-example groups. Circles and squares indicate the average test score for each group. The error bars show the standard error.}
\label{fig:ere_with_var_data}
\Description{The vertical axis represents post-test scores, ranging from 40.0 to 80.0.}
\end{figure}

We find that these LM-based SEEs successfully replicate the Variability Effect. We plot the post-test scores for all participants in Figure \ref{fig:var_results}. High variability in problems significantly enhances performance in worked-example conditions, whereas it has no effect in practice conditions. Additionally, our data from these SEEs also replicates the Expertise Reversal Effect (see Figure \ref{fig:ere_with_var_data}), suggesting that the LM judgments are consistent. As discussed in Section \ref{sec:ere_results}, despite the LM potentially being trained on datasets that include descriptions of previous studies, our evaluations use new instructional materials from a different domain than prior studies and do not reference these studies. This demonstrates the LM's capability to act as a reliable evaluator. 

%% file: sections/4_optimization.tex
\section{Instruction optimization}
We showed that LMs have relevant educational knowledge when used via SEEs, which suggests SEEs can be used to craft better instructional materials.

\subsection{The Instruction Optimization algorithm }

We introduce the Instruction Optimization algorithm (see Alg.~\ref{alg:optimizer}), which iteratively refines instructional materials by using an evaluator LM ($M_e$) and an optimizer LM ($M_o$). In this algorithm, $M_o$ generates new instructions based on prior instructions and resulting learning outcomes (e.g., post-test scores) of a particular student, and $M_e$ evaluates these new instructions by predicting the student's learning outcomes. 

More precisely, the input of the algorithm are $M_e$, $M_o$, an initial text instruction \(c_0\), a text description of a student persona, an evaluation task description, an optimization task description, and $T$ test questions $x_1,...,x_T$. We use an SEE (Alg.~\ref{alg:evaluator}) to assess the quality of \(c_0\) and receive an initial post-test score \(r_0\). We store the instruction-score pair $(c_0, r_0)$ in $mem$. Following the prompt design strategy in \cite{yang2023large}, we construct the initial optimization prompt $p_0$ (see an example in Figure \ref{fig:optimization_prompt}) that includes the student persona, the instruction-score pair(s) in $mem$, and the optimization task description for the LM (which is to generate a new instruction to further increase the post-test score of the student). In each optimization step $n$, we give $p_{n-1}$ as input to $M_o$, which generates a new instruction. We repeat this $K$ times and generate $K$ new instructions for each optimization step. We evaluate each of the $K$ instructions using Alg.~\ref{alg:evaluator} and store the resulting instruction-score pairs in a temporary memory $mem'$. We then add instruction-score pairs in $mem'$ to $mem$. We sort instruction-score pairs in $mem$ by their scores in ascending order and remove the instructions with the lowest scores from $mem$ until $mem$ meets the maximum length requirement (due to the LM's prompt length limit). We update the prompt using the updated $mem$ and use the new prompt for the next iteration. 

\begin{algorithm}[h!]
 \caption{Instruction Optimization}
  \begin{algorithmic}
    \State \textbf{Input:} Evaluator $M_e$, Optimizer $M_o$, \\ Initial instruction $c_0$, \textit{student\_persona}, \textit{eval\_task}, \\
    \textit{optimization\_task}, test questions $x_1,...,x_T$
    \State Evaluate $c_0$ using Algorithm \ref{alg:evaluator} and receive score $r_0$
    \State Set $\mathcal D \gets \{(c_0,r_0)\}$
    \State Set $mem \gets [(c_0,r_0)]$
    \State Set $p_0 \gets \textit{student\_persona} || mem || \textit{optimization\_task}$ 
    \For{$n = 1,2,\dots N$}
        \State Set $mem' \gets [ ]$
        \For{$k = 1,2,\dots K$}
            \State Generate $c_{(n-1)K + k}$ by feeding $p_{n-1}$ into $M_o$
            \State Evaluate $c_{(n-1)K + k}$ using Algorithm \ref{alg:evaluator} and receive score $r_{(n-1)K + k}$
            \State Append $(c_{(n-1)K + k},r_{(n-1)K + k})$ to $mem'$
            \State Set $\mathcal D \gets \mathcal D \cup \{(c_{(n-1)K + k},r_{(n-1)K + k})\}$
        \EndFor
        \State Set $mem \gets mem \cup mem'$
        \State Sort $mem$ by scores in ascending order
        \State Set $mem \gets mem[:\textrm{max length}]$
        \State Set $p_n \gets \textit{student\_persona} || mem || \textit{optimization\_task}$
    \EndFor       
    \State \Return $\mathcal{D}$
  \end{algorithmic}
  \label{alg:optimizer}
  \end{algorithm}

\begin{figure*}[ht]
\noindent\fbox{
\parbox{0.975\textwidth}{

\color{tableauBlue}
Here is an 8th-grade student with the following skill levels (each skill is rated on a scale from 1 to 5, where higher numbers indicate more proficiency):
\begin{enumerate}
    \item Being able to set up systems of equations given a word problem: 1
    \item Being able to solve systems of equations: 1
\end{enumerate}

\color{tableauOrange}

\quad 

I have some worksheets along with the student's test scores after receiving the worksheets. The worksheets are arranged in ascending order based on their scores, where higher scores indicate better quality.

\quad

Worksheet: 

You need to study a problem and its solution. Here's the problem: 
A brownie recipe is asking for 350 grams of sugar, and a pound cake recipe requires 270 more grams of sugar than a brownie recipe. How many grams of sugar are needed for the pound cake? Here's its solution: Step 1: Identify the amount of sugar needed for the brownie recipe, which is 350 grams. Step 2: Understand that the pound cake recipe requires 270 more grams of sugar than the brownie recipe. Step 3: Add the additional 270 grams of sugar to the 350 grams required for the brownie recipe. Step 4: The total amount of sugar needed for the pound cake recipe is 350 grams + 270 grams = 620 grams.

Test score: 20

\quad

Worksheet:

\{worksheet2\}

Test score: 34

\quad

Worksheet:

\{worksheet3\}

Test score: 66

\quad   

(...more worksheets and scores...)

\quad

\color{tableauGreen}
Generate a new worksheet to further increase the test score of the student. You will be evaluated based on this score function:

```python

\{utility\_string\}

'''

The new worksheet should begin with <WORKSHEET> and end with </WORKSHEET>.

}  
}

\caption{The optimization prompt for generating a new instruction (e.g., a worksheet) based on prior instructions and post-test scores for a given student. The \textcolor{tableauBlue}{blue} text is the student persona. The \textcolor{tableauOrange}{orange} text contains the previous worksheet-score pairs, sorted in ascending order. The \textcolor{tableauGreen}{green} text describes the optimization task for the LM. The score function is a Python program that illustrates how post-test scores are computed without revealing the content of the post-test.}
\label{fig:optimization_prompt}
\Description{The optimization prompt for generating a new instruction (e.g., a worksheet) based on prior instructions and post-test scores for a given student. The \textcolor{tableauBlue}{blue} text is the student persona. The \textcolor{tableauOrange}{orange} text contains the previous worksheet-score pairs, sorted in ascending order. The \textcolor{tableauGreen}{green} text describes the optimization task for the LM. The score function is a Python program that illustrates how post-test scores are computed without revealing the content of the post-test.}
\end{figure*}

\subsection{Optimizing math worksheets}

\begin{figure}[h!]
    \centering    \includegraphics[width=0.825\columnwidth]{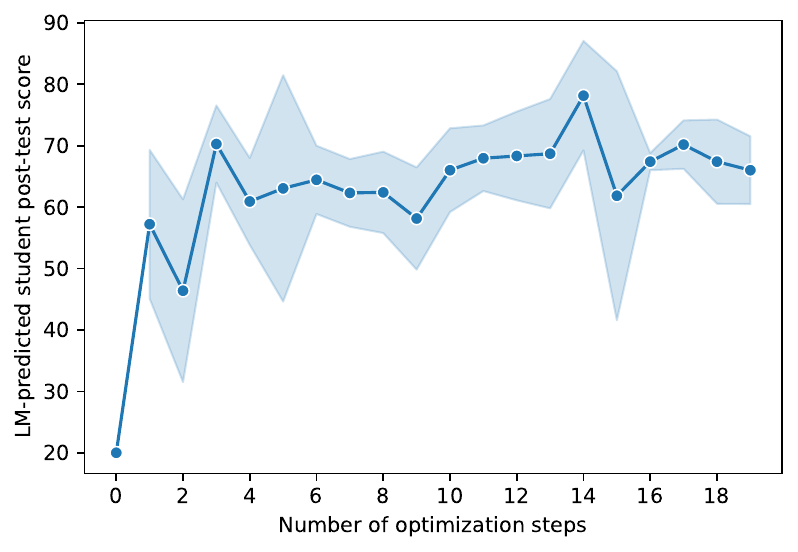}
    \caption{LMs can iteratively improve math word problem worksheets using their own judgments as the reward function. Each dot represents the average post-test score of $3$ worksheets generated at each optimization step, with error bars indicating the standard deviation.}
    \label{fig:optimization_results}
    \Description{This plot shows a line graph depicting the relationship between the number of optimization steps and the LM-predicted student post-test scores. The horizontal axis of the graph represents the number of optimization steps, which range from 0 to 18. The vertical axis measures the LM-predicted student post-test scores, ranging from 20 to 90. The line graph starts at a lower score at zero optimization steps and generally trends upward as the number of optimization steps increases.}
\end{figure}

\begin{figure*}[ht]
    \centering    \includegraphics[width=\textwidth]{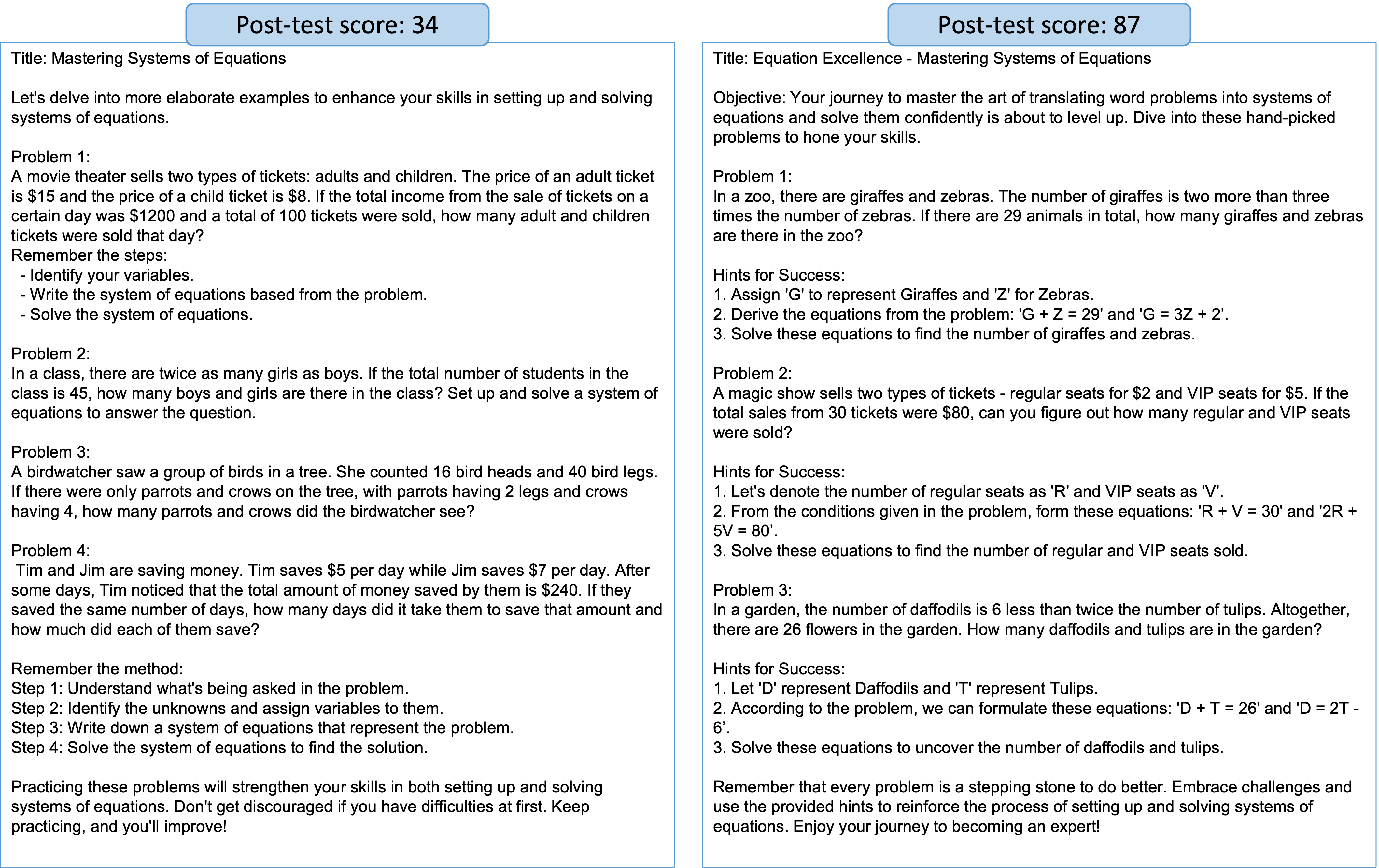}
    \caption{Two examples of worksheets generated by the optimizer LM. As predicted by the evaluator LM, the left one leads to a post-test score of $34$, and the right one gives a post-test score of $87$.}
    \label{fig:bad_vs_good_worksheet}
    \Description{Two examples of worksheets generated by the optimizer LM.}
\end{figure*}

We demonstrate the effectiveness of our Instruction Optimization algorithm (Alg.~\ref{alg:optimizer}) on optimizing math word problem worksheets. Starting with an initial worksheet that yields a low post-test score (see the first worksheet-score pair in Figure \ref{fig:optimization_prompt}), we demonstrate that the optimizer LM (GPT-4) can improve the post-test score of the generated worksheets until convergence (see Figure \ref{fig:optimization_results}). Figure \ref{fig:bad_vs_good_worksheet} shows two examples of LM-generated worksheets. 

We describe the implementation details below. We use gpt-3.5-turbo-16k as $M_e$ and gpt-4 as $M_o$. We use temperature $= 0$ for gpt-3.5-turbo-16k, and we use temperature $=1$ for gpt-4 to encourage more diverse generations of new worksheets. We generate $K=3$ worksheets per optimization step. We run a total of $N=19$ steps. When we evaluate each worksheet using Alg.~\ref{alg:evaluator}, we run three independent evaluations and take the average post-test score across the three evaluations as the final post-test score for each worksheet, which allows $M_o$ to get more stable reward signals. We randomly select $T=6$ problems from the \textsc{Algebra} dataset \cite{he2023solving} as the post-test. The post-test is identical in all evaluations. The max length limit for $mem$ is set to be 8 worksheet-score pairs. We use a fixed student persona, where the student's initial skill levels are low (see the \textcolor{tableauBlue}{blue} text in Figure \ref{fig:optimization_prompt}), to simulate a personalized instructional design process. In the optimization task description, we use a Python program to show how post-test scores are computed without showing specific problems on the post-test (see Figure \ref{fig:utility_string} in Appendix).

\subsection{Human evaluation of worksheets}

We conducted a human teacher evaluation on worksheets generated by the LM, involving 95 participants from Prolific. All participants met the following criteria:
\vspace{-0.4cm}
\begin{enumerate}
    \item The participant has a background in teaching.
    \item The participant is based in the U.S.
    \item The participant is fluent in English.
    \item The participant has an approval rate over $95\%$ on Prolific.
\end{enumerate}
\vspace{-0.3cm}
All participants provided informed consent before participation following an approved institutional review board (IRB) protocol\footnote{The risks associated with this study are minimal. Participants were told that the study data would be stored securely, minimizing the risk of confidentiality breach. Their individual privacy is maintained during the research and in all published and written data resulting from the study.}. We paid participants at a rate of 12 USD per hour. 

\begin{figure*}
    \centering
    \includegraphics[width=0.9\textwidth]{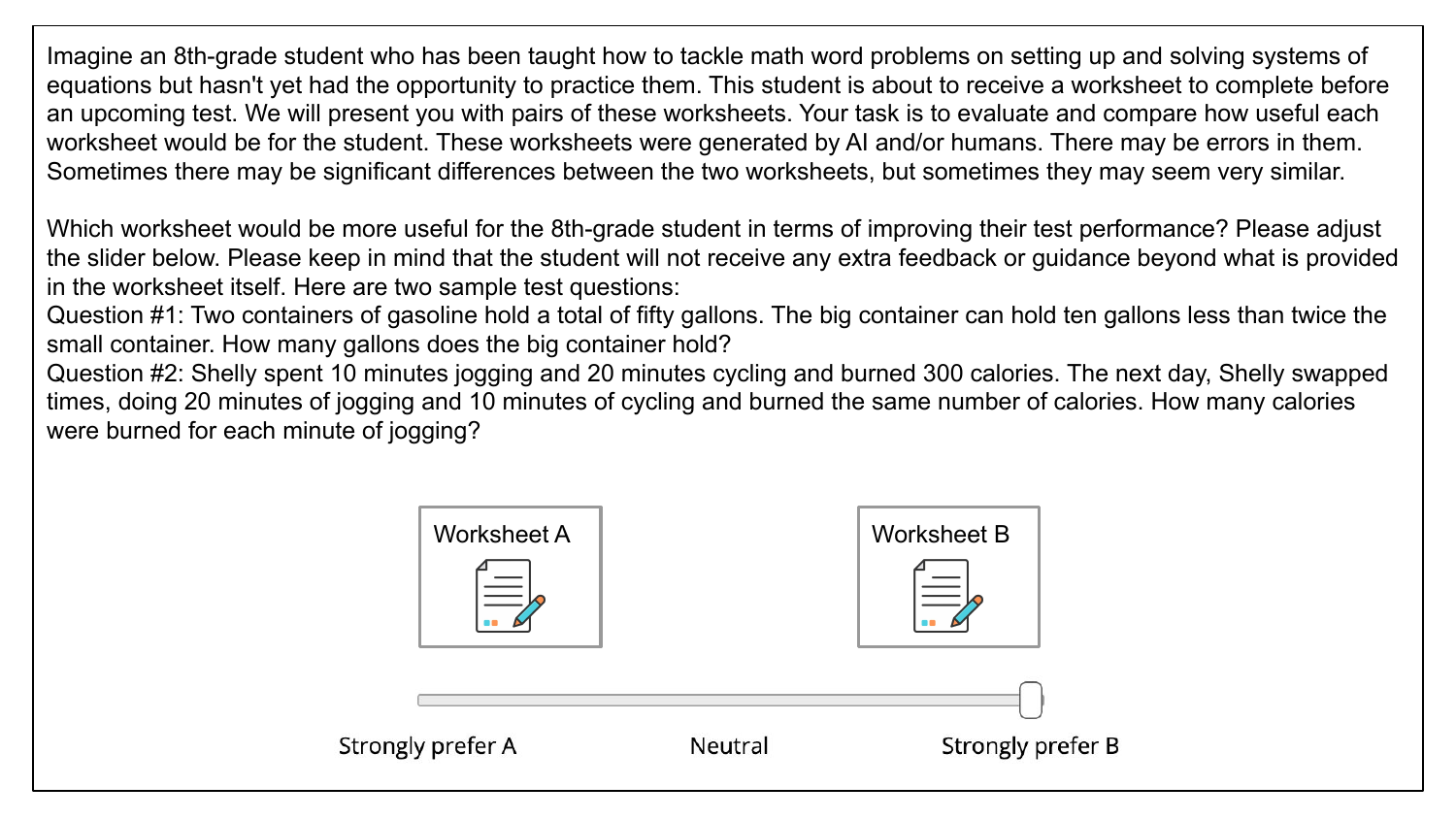}
    \caption{The evaluation task description for teachers. In each evaluation, we ask the teacher to compare two worksheets in terms of their effectiveness in improving the student’s test performance, and we show the teacher two sample test questions. Teachers can indicate their preference by adjusting the slider.}
    \label{fig:human_teacher_evaluation_task}
    \Description{This outlines a human evaluation task related to assessing the utility of math worksheets for an 8th-grade student.}
\end{figure*}

We asked participants to do pairwise comparisons where they need to indicate their preferences between two worksheets (worksheet A and B) using a continuous slider scale, with a rating of $-1$ assigned for a strong preference towards the first worksheet, a rating of $1$ for a strong preference towards the second worksheet, and a rating of $0$ for no preference (see the task description in Figure \ref{fig:human_teacher_evaluation_task}).

We evaluated a total of 10 worksheets, of which one was the initial worksheet we provided to the optimizer LM, and the rest were randomly selected from LM-generated worksheets. There are 45 unique pairs of worksheets, and we randomly assigned $4$ or $5$ pairs of worksheets to each participant. We also assigned an additional catch-trial question after the first three pairs of worksheets to filter participants who did not pay attention. We collected a total of $465$ ratings from $n=95$ participants, and each pair of worksheets has at least $7$ ratings. After filtering out participants who failed the catch trial, we have a total of $270$ ratings from $56$ participants. We calculate the human preference score for each worksheet $c$, $hps(c)$, by accumulating the negative ratings when a worksheet was less preferred and the positive ratings when it was more preferred, normalized by the total number of times each worksheet was compared:
\vspace{-0.02cm}
\begin{equation}
    hps(c) = \frac{\sum_{(a,b) \in \mathcal{S}_A^c} -R(a, b) + \sum_{(a,b) \in \mathcal{S}_B^c} R(a, b)}{|\mathcal{S}_A^c| + |\mathcal{S}_B^c|},
\end{equation}
\vspace{-0.5cm} 

where $\mathcal{S}_A^c$ denotes all the pairs of worksheets where $c$ appears as the first worksheet (which means we need to negate the rating to indicate how much $c$ is preferred over the other worksheet), and $\mathcal{S}_B^c$ denotes all the pairs of worksheets where $c$ appears as the second worksheet. $|\mathcal{S}_A^c|$ and $|\mathcal{S}_B^c|$ denote the size of the each set.

\begin{figure}[ht]
    \centering    
    \includegraphics[width=\columnwidth]{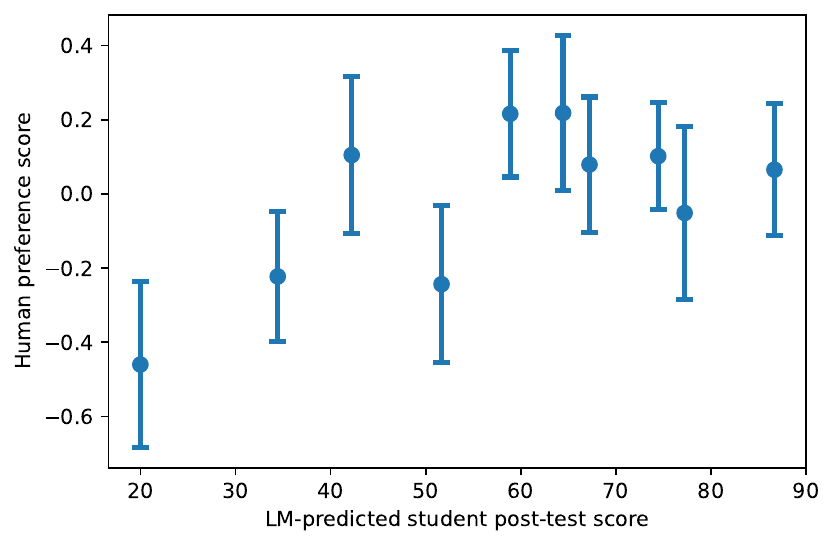}
    \caption{Evaluations by human teachers ($n=56$) show a significant correlation between the LM judgments and human teacher preferences (Pearson $r=0.661, p<0.05$). The error bars are bootstrapped 95\% confidence intervals.}
    \label{fig:human_preference_vs_lm_judgments}
    \Description{This is a scatter plot comparing LM-predicted student post-test scores with human preference scores.}
\end{figure}

We calculate the Pearson correlation between post-test scores predicted by the LM and the human preference scores for all worksheets and find a significant correlation ($r=0.661, p<0.05$) between the LM judgments and human teacher preferences (see Figure \ref{fig:human_preference_vs_lm_judgments}). However, human teachers sometimes cannot distinguish worksheets that LMs identify as distinct, particularly those with predicted post-test scores ranging from 60 to 90.

%% file: sections/5_conclusion.tex
\section{Limitations}
There are several limitations to our work. First, the LMs we used were likely trained on data containing descriptions of experiments similar to those we replicated. Therefore, we developed instructional materials in a domain different from previous studies.

Second, one potential concern is that LMs do not consistently replicate established educational results. Our experiments focused solely on mathematics and only tested the performance of GPT-3.5 and GPT-4 with a limited number of straightforward prompts. Exploring all available LMs would have been prohibitively costly, so we chose GPT-3.5 and GPT-4 due to their popularity and robust performance. We used GPT-3.5 as the evaluator LM because it is more cost-effective for handling long prompts. GPT-4 served as our optimizer LM, chosen for its capability in critical assessment, a feature typically reliable only in the most advanced models.  Our goal was to demonstrate the potential of LMs in these tasks, not to claim these LM choices as the definitive best.

Third, there are discrepancies between the evaluations made by LMs and those by human teachers, who sometimes fail to perceive differences that LMs identify. These discrepancies raise concerns about the models' decision-making processes and the potential impact of biases in their training data. Furthermore, our evaluations are limited to the perspective of teachers. However, ultimately, the effectiveness of the educational materials depends on how students engage with them. While LMs could aid in accelerating the design process by providing preliminary evaluations, they should not be seen as replacements for direct evaluations involving human students.

\section{Conclusion}
We show that LMs can act as evaluators on a couple of well-known findings for pedagogical effectiveness, including results that require understanding that different students can be impacted differently by different content. Even if the LM succeeds at these judgments due to reading relevant studies, it is still remarkable for two reasons: 1) Nothing in the evaluations directly cued those studies, and 2) We developed new instructional materials in a domain different from the ones explored in prior studies. This shows that the LM can read and synthesize educational science reliably into useful evaluators. In addition, LM evaluators correlated positively with expert humans in their evaluations of the effectiveness of new content.

However, there may be other known educational findings that LM evaluators do not replicate, and human teachers sometimes cannot distinguish instructional materials that LMs perceive as different and may disagree with LMs on what is optimal. Therefore, the LM may be useful as a coarse evaluator to help speed up content design as a human augmentor but not to replace human evaluators or empirical studies. An interesting open question is whether we can extend LM-based simulated educational experts to multi-modal instructional input. 

%% file: sections/appendix.tex
\section{Modeling cognitive processes of learning dynamics}
\label{appendix:student_simulator_challenge}

\begin{figure}[H]
    \centering
    \includegraphics[width=0.9\columnwidth]{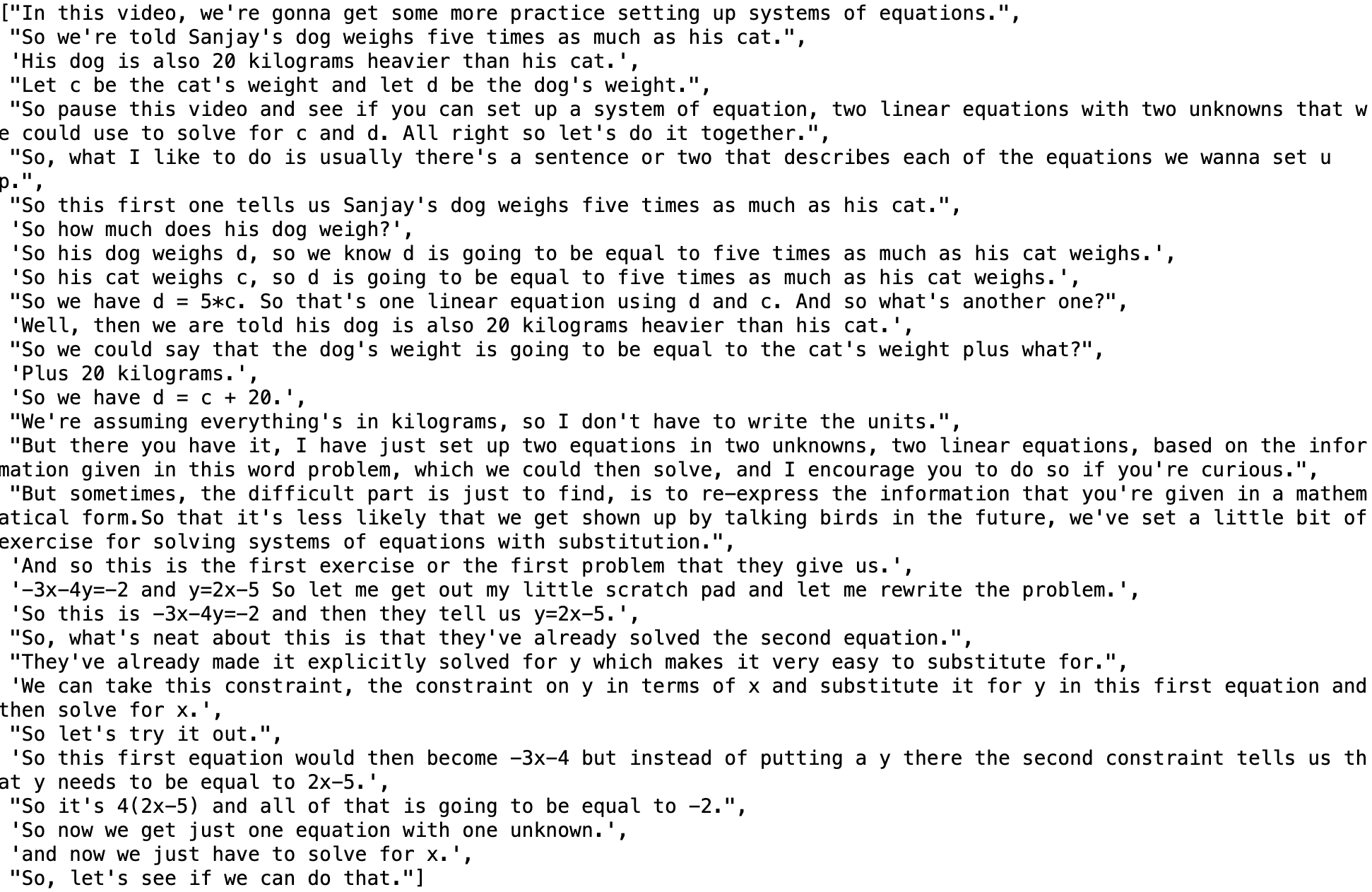}
    \caption{The first 30 sentences in the video transcript.}
    \label{fig:video_transcript}
    \Description{see caption.}
\end{figure}

We simulate a scenario where LMs play the role of students learning from scratch, engaging with a Khan Academy video transcript to understand how to set up and solve systems of equations (see our prompt in Figure \ref{fig:prompt_student_simulator}). To test the models' understanding, we give them a quiz with 14 math word problems related to the topic, evaluating their performance after they have read every few sentences of the transcript (see the first 30 sentences of the transcript in Figure \ref{fig:video_transcript}). 

\begin{figure}[H]
    \centering    \includegraphics[width=0.45\columnwidth]{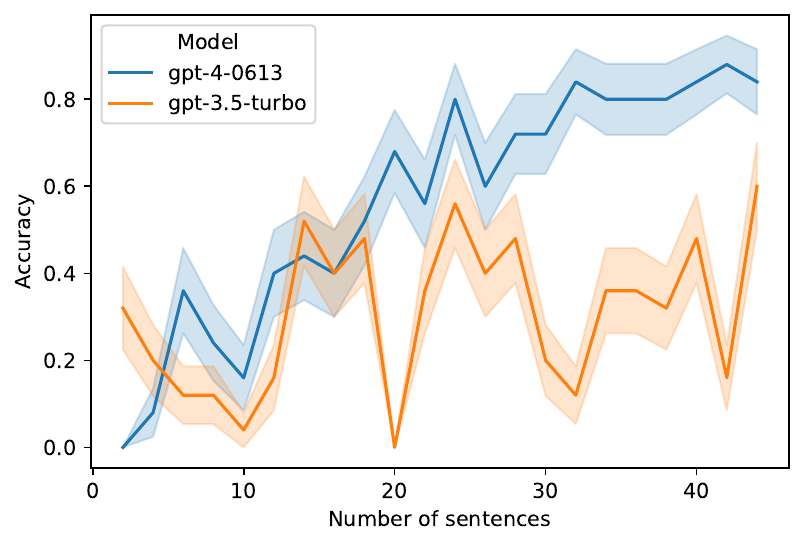}
    \caption{Simulating the dynamics of learning with language models (LMs) poses a challenge. Unlike human learners, LMs can shift from having no knowledge to being proficient after only a brief interaction, such as reading just a few sentences from a lecture.}
    \label{fig:student_simulator_results}
    \Description{shows a comparison of accuracy rates between two models, GPT-4-0613 and GPT-3.5 Turbo, across a range of sentence counts. The horizontal axis of the graph indicates the number of sentences, ranging from 0 to 40. The vertical axis measures accuracy, ranging from 0.0 to 0.8.}
\end{figure}

We find that LMs would often transition from a lack of algebraic knowledge to proficiency after a minimal interaction (e.g., like reading just a few sentences from a lecture), which is not a realistic representation of how humans learn. We plot LMs' performance accuracy on the quiz over the number of sentences LMs have read in Figure \ref{fig:student_simulator_results}. GPT-4, despite being asked to act as a novice, solves about $40 \%$ of the problems on the quiz after encountering the first 6 sentences of the video transcript.

\begin{figure}[H]
\noindent\fbox{
\parbox{0.975\textwidth}{

% \color{cbBlue}
\color{tableauBlue}
You are a student who is initially ignorant. 
You are learning by watching a video. Here's the video:  

\quad  

\color{black}

In this video, we're gonna get some more practice setting up systems of equations.

So we're told Sanjay's dog weighs five times as much as his cat.

His dog is also 20 kilograms heavier than his cat.

\quad

% \color{cbOrange}
\color{tableauBlue}

Let's stop the video. Remember, you've only been taught what was shown in the video. It typically takes you 5-10 practices to learn a new skill, and you often need to rewatch a lecture and do practice problems. Remember you have done 0 examples, and that is less than 5!
Can you apply what you've just learned to solve the following problem? If you do not know how to solve it, don't try to guess and ask to keep watching the video. If you do know how to solve it, then place the final answer (a number) in square brackets. Here's the problem:

% \quad 

Alyssa is twelve years older than her sister, Bethany. The sum of their ages is forty-four. Find Alyssa's age.

% \quad

}  
}
\caption{The \textcolor{black}{black} text is the video transcript. In this example, we show the LM the first three sentences of the transcript.}
\label{fig:prompt_student_simulator}
\Description{see caption.}
\end{figure}

\section{Prompts}
Prompts used in this work are publicly available at \href{https://github.com/StanfordAI4HI/ed-expert-simulator}{https://github.com/StanfordAI4HI/ed-expert-simulator}.

\begin{figure}[H]
\noindent\fbox{
\parbox{0.975\columnwidth}{

\color{tableauBlue}
Here is an 8th-grade student with the following skill levels (each skill is rated on a scale from 1 to 5):
\begin{enumerate}
    \item Being able to set up systems of equations given a word problem: \{level1\}
    \item Being able to solve systems of equations: \{level2\}
\end{enumerate}

\color{tableauOrange}

\quad 

Now the student is asked to work on the following problem on a test: 

\quad   

\{problem\}

\quad

\color{tableauGreen}
Given the student's initial skill levels, what's the probability that the student can solve the problem correctly? Explain your reasoning and give a single number between 0 and 100 in square brackets.

}  
}
\caption{The prompt for predicting the pre-test score in a simulated expert evaluation. The \textcolor{tableauBlue}{blue} text is the student persona. The \textcolor{tableauOrange}{orange} text is the test. The \textcolor{tableauGreen}{green} text describes the evaluation task for the LM.}
\label{fig:see_pre_test_prompt}
\Description{see caption.}
\end{figure}

% \section{Optimizing math word problem worksheets}

\begin{figure}[H]
    \centering
    \includegraphics[width=0.82\columnwidth]{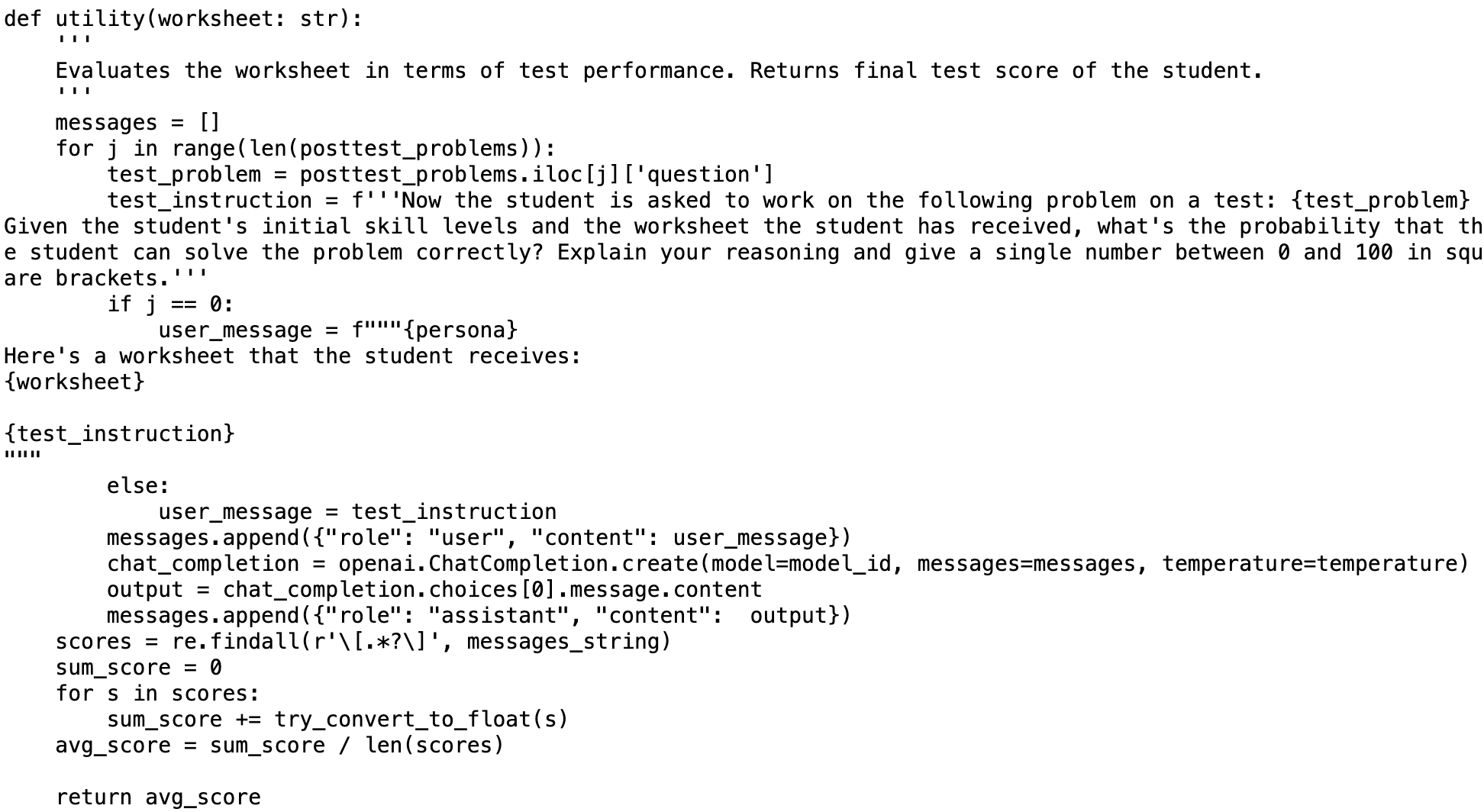}
    \caption{The utility string is a Python program that shows how the post-test score of a given worksheet is computed.}
    \label{fig:utility_string}
    \Description{see caption.}
\end{figure}